\pdfoutput=1
\PassOptionsToPackage{dvipsnames}{xcolor}
\documentclass[11pt]{article}

\usepackage[final]{acl}

\usepackage{times}
\usepackage{latexsym}
\usepackage{todonotes}
\usepackage[T1]{fontenc}

\usepackage[utf8]{inputenc}

\usepackage{microtype}

\usepackage{inconsolata}

\usepackage{graphicx}

\usepackage{amssymb}
\usepackage{amsmath}

\usepackage{booktabs}
\usepackage{multirow}
\usepackage{multicol}
\usepackage{float}
\usepackage{mathtools}
\usepackage{bbold}
\usepackage{placeins} 
\usepackage{stfloats}

\usepackage{fancyvrb}
\usepackage{xcolor}

\usepackage{pifont}
\newcommand{\cmark}{{\color{ForestGreen}\ding{51}}}%
\newcommand{\xmark}{{\color{BrickRed}\ding{55}}}%

\title{Learning to Translate Ambiguous Terminology by Preference Optimization on Post-Edits}

\author{Nathaniel Berger$^{\ast}$, Johannes Eschbach-Dymanus$^\ddag$, Miriam Exel$^\ddag$, Matthias Huck$^\ddag$ \and Stefan Riezler$^{\dagger,\ast}$\\ 
  $^{\ast}$Computational Linguistics \& $^\dagger$IWR, Heidelberg University, Germany \\
  $^\ddag$SAP SE, Dietmar-Hopp-Allee 16, 69190 Walldorf, Germany\\
  {\tt \{berger, riezler\}@cl.uni-heidelberg.de}\\  
  {\tt \{johannes.eschbach-dymanus, miriam.exel, matthias.huck\}@sap.com}}

\begin{document}
\maketitle
\begin{abstract}

In real world translation scenarios, terminology is rarely one-to-one. Instead, multiple valid translations may appear in a terminology dictionary, but correctness of a translation depends on corporate style guides and context. This can be challenging for neural machine translation (NMT) systems. Luckily, in a corporate context,  many examples of human post-edits of valid but incorrect terminology exist. The goal of this work is to learn how to disambiguate our terminology based on these corrections. Our approach is based on preference optimization, using the term post-edit as the knowledge to be preferred. While previous work had to rely on unambiguous translation dictionaries to set hard constraints during decoding, or to add soft constraints in the input, our framework requires neither one-to-one dictionaries nor human intervention at decoding time. We report results on English-German post-edited data and find that the optimal combination of supervised fine-tuning and preference optimization, with both term-specific and full sequence objectives, yields statistically significant improvements in term accuracy over a strong NMT baseline without significant losses in COMET score. 
Additionally, we release test sets from our post-edited data and terminology dictionary. 

\end{abstract}

\section{Introduction}

\begin{figure}[t]
\centering
\includegraphics[width=0.5\textwidth]{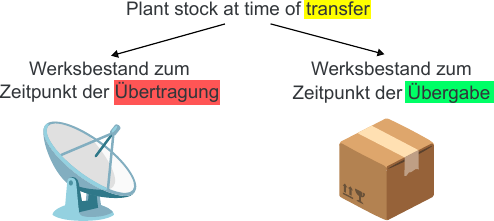}
\caption{The term 'transfer' is a highly ambiguous term in our dictionary with 27 possible translations. Transfer could be translated to "Übertragung" which would refer to a transfer of data or a broadcast. On the other hand, "Übergabe" could mean a transfer of physical goods. In this case, the source sentence is likely referring to physical goods with "Plant stock".}
\label{fig:example}
\end{figure}

In business scenarios, accurate terminology translation is critical to ensure that the translated text is understood as intended. Ambiguous terminology can make this more difficult. Take, for example, the term 'transfer' (Figure \ref{fig:example}). In German, our terminology dictionary specifies 27 possible translations of the term 'transfer', depending on the context and part of speech. A 'transfer' could be a delivery, in which case it is a 'Übergabe' or 'Warenüberführung', whereas on the other hand, it could be a transfer of data, which makes it a 'Übertragung'. If you are calling a company on the phone and they have to transfer you to another department, it is a 'Weiterleitung'. Working with a logistics company, any of these could be possible translations of 'transfer'. 
Neural machine translation (NMT) is quite capable of using sentence level context to disambiguate terminology, but it is not perfect and frequently human translators are required to intervene and correct machine translations to create post-edited translations. 
However, even though the professional translators who perform post-editing are knowledgeable about how to translate terms depending on their usage, which term translation they use varies from editor to editor.
Moreover, if NMT is provided to end users, it faces the challenge that users may not know the correct term translations or even speak one of the languages. This makes approaches that require annotating source terms with the desired target translation less tenable. Automated solutions are possible if an unambiguous dictionary is available, or else a system could detect when the source text contains a term and annotate it with all possible term translations, but it still is the responsibility of the NMT system to determine the correct translation.

\begin{figure*}[t]\includegraphics[width=\textwidth]{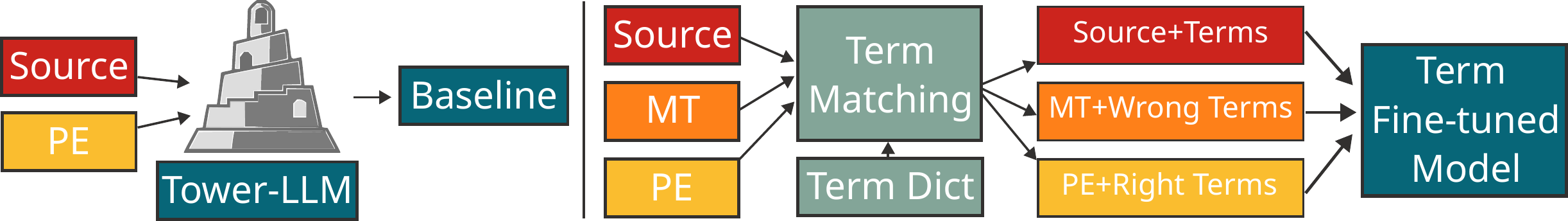}
    \caption{In order to train terminology corrected models, we begin by performing continued pre-training on Tower-7B base on in-house post-edits to serve as our baseline. We then match our terminology in post-edits and MT to find MT with incorrect but valid terms and post-edits with corrected terms. This is used to train our baseline for term fine-tuned models.}
\label{fig:process_flowchart}
\end{figure*}

This work is a step towards an NMT system that is more capable of producing correct term translations without the need for a one-to-one terminology dictionary, or for human intervention at decoding time. The central idea is to to learn terminology translation from statistical information about context and editor preferences that is encoded in post-edits. We show that a combination of supervised fine-tuning and preference optimization (PO) for knowledge editing on (post-edit, machine translation) pairs \citep{rozner-etal-2024-knowledge,berger-etal-2024-post} with term-specific and sequence level objectives, the capabilities of NMT to disambiguate depending on context knowledge can be significantly improved by exploiting fine-grained discriminative information about preferred and dispreferred term translations. 
We create a variant of IPO \citep{pmlr-v238-gheshlaghi-azar24a} and of supervised fine-tuning (SFT) that masks non-term tokens. We then create combinations of masked and unmasked losses to investigate what benefits ambiguous term translation. Our experimental results show that PO learns a margin between preferred and dispreferred sequences, with the effect that semantically similar term translations are separated and the correct terminology variant is selected. Evaluation scores on English-German post-edited data show that the optimal combination of SFT and PO, where we combine a masked objective to target terminology with training on full sequences, yields statistically significant improvements in term accuracy over a strong NMT baseline (Tower Base LLM 7B) pre-trained on in-domain data, without significant losses in COMET score. Translation with foundational LLMs (GPT 4.1) yields competitive COMET scores, but cannot exploit ambiguous term dictionaries profitably.

A further contribution of our work is a release of terminology annotated test sets, containing source text, machine translations, and post-edits. Half of each test set contains ambiguous terms where the MT uses an incorrect term variant and the PE corrects term in the MT.

\section{Related Work}

Previous works on terminology in translation have focused on adding either hard constraints on decoding, such as in constrained beam-search (\citet{hokamp-liu-2017-lexically}, \citet{post-vilar-2018-fast}), or adding a soft constraint by providing the source and target term pair as an additional model input (\citet{dinu-etal-2019-training}, \citet{exel-etal-2020-terminology}). Adding the terminology translation as an extra input has been expanded upon with the rise of LLMs and in-context learning to prompt-based methods \citep{moslem-etal-2023-adaptive}. The latter approach was the most popular in the latest iteration of the WMT shared task on terminology translation \citep{semenov-etal-2023-findings}, with all six of the evaluated systems using terms as an additional input and a few adding either a post-processing step or modifying decoding strategies to further enforce term usage. 
With both hard-constraints during decoding or soft-constraints in the input, one most know beforehand what the correct term translation is. For term dictionaries that contain a one-to-one term mapping, this is no problem. However,  when the term dictionary contains a one-to-many mapping, the decision of what term translation is correct is offloaded from the system to a human annotator or translator.

Work has been done to add more advanced fulfillment criteria to the constrained beam search that allow multiple variants of a term to fulfill the constraint. For example \citet{hauhio-friberg-2024-mitra}, use finite state automata to check for constraint fulfillment and construct automata that recognize all acceptable term inflections. However, in order to only accept a specific term translation, one would again have to manually specify what the desired translation is.

Our approach is based on preference optimization for text generation models \citep{ChristianoETAL:17,rafailov2023direct,pmlr-v238-gheshlaghi-azar24a,icml24-contrastive-preference} applied to performing knowledge editing \citep{rozner-etal-2024-knowledge} on our baseline model. We begin with a baseline model that is already quite capable of producing valid terms that exist in our term dictionary. From  the possible valid terms, we would like to produce exactly \textit{the} correct term and not \textit{an} extant term. 

\section{Methods}

Our task varies from the WMT terminology translation task in the following ways: first, the terminology dictionaries used to evaluate the shared task were created by aligning source and reference texts and having human annotators correct false alignments or incorrect terminology. Second, the segment level term dictionaries were provided to the NMT system at evaluation time. Segment level term dictionaries created by alignment ensure that, for a term in a given segment, there is only one correct translation. 

We focus on applying an existing terminology dictionary with ambiguous terms, containing multiple translations for a given source term. 
Additionally, we do not provide the term dictionary as input to the model. 
In order to tackle our problem of ambiguous terminology, we need to find examples of terms being incorrectly used and then corrected, and training objectives that make use of the pair of (correct, incorrect) terminology. An overview of our process is illustrated in Figure \ref{fig:process_flowchart}.

\subsection{Terminology Matching}
\label{terminology_matching}
To find incorrect terms in our machine translations and their corrections in our post-edits, we use fuzzy matching. We use the RapidFuzz library\footnote{https://github.com/rapidfuzz/RapidFuzz} to calculate fuzzy matching ratios and align the terms in the dictionary and text. This is done with the function \texttt{partial\_ratio\_alignment}, which returns a span in the text and a score. The fuzzy match score is calculated as 1 minus the normalized Levenshtein distance, with substitution cost of 2. The score threshold is set to 0.95 to allow minor variations.

First, we check if the source text contains a valid term. If it does, we then look for all possible term translations in both the MT and PE. If both the MT and PE contain terms, we then make sure that they don't contain the same term. Because some term translations overlap (both 'transfer' and 'transferieren' would match if the latter were contained in the text), we may have a set of matches for MT or PE. We accept the term matches if the intersection of the MT and PE term match sets is empty.

As some terms are subsequences of other terms, i.e. "Überführung" is wholly contained within "Warenüberführung", we remove overlapping terms by checking if one matched term is contained within another matched term. If that is the case, then we take the larger of the two matches.

\subsection{Training Objectives}

Our training objective modifies the dCPO loss of \citet{berger-etal-2024-post} for preference optimization on terminology-containing post-edit pairs. The dCPO loss is a direct preference optimization loss based on the alternative formulation of preference optimization by \citet{pmlr-v238-gheshlaghi-azar24a}, which introduced the IPO loss, combined with the insight of \citet{icml24-contrastive-preference} that adding an SFT term improved the ability of the language model to learn the reference translations. It is defined as
\begin{multline}
\mathcal{L}_{dCPO}(y_w, y_l, x) = - \log(\pi^*(y_w|x)) \,+\, \\
\left(\!\left(\!\log\!\left(\!\frac{\pi^*(y_w|x)}{\pi_{ref}(y_w|x)}\!\right) \!- \!\log\!\left(\!\frac{\pi^*(y_l|x)}{\pi_{ref}(y_l|x)}\!\right)\!\right) \!-\!\frac{1}{2\beta}\!\right)^2
\end{multline}
where $\pi^*$ is the model currently being trained, $\pi_{ref}$ is the initial model, $y_w$ is the preferred sequence, and $y_l$ is the dis-preferred sequence. $\pi(y|x)$ is calculated as $$\sqrt[|y|]{\Pi_{i=0}^{|y|}\pi(y_i|y_{<i},x)}$$ which is the geometric mean of target token-level probabilities, the log of which is proportional to the arithmetic mean of log-probabilties as described in \citet{berger-etal-2024-post}. The $\pi_{ref}$ in the denominator serves as a KL-divergence regularizer on the loss so that baseline performance is not lost. However, because we are focused on fixing ambiguous terms in our baseline model, we do not want to regularize towards this behavior and therefore remove it, similar to \citet{icml24-contrastive-preference}.

During training, we noticed that the SFT term contained in the dCPO loss can easily be overwhelmed by the IPO term when the distance set in the IPO term is much larger than the initial distance between sequences in log-probability space. Because IPO is based on a squared-error loss, gradients grow multiplicatively with the error. This is problematic if the  baseline model has already fit the post-edits well, meaning the SFT term's gradient has a fairly small norm in comparison to the IPO term.
To ameliorate this problem, we introduce two modifications. First, we replace the squared-error component of the loss with smooth-l1. 
\begin{multline}
    sl_1(x,y) = 
    \begin{cases}
        0.5(x-y)^2 & \text{if} \,|x-y| < 1\\
        |x-y|-0.5 & \text{otherwise}
    \end{cases}
\end{multline}
This removes the multiplicative growth of the gradient with regard to the input, to reduce the exploding gradient problem. Thus our preference optimization loss becomes 
\begin{multline}
    \mathcal{L}_{PO}(x,y_w,y_l) = \\ sl_1\left(\log(\pi^*(y_w|x)) - \log(\!\pi^*(y_l|x)), \tfrac{1}{2\beta}\right)
    \label{eq:po}
\end{multline}

Our second addition is to simply add a weight $\alpha$ on the SFT loss term to attempt balancing it with the PO loss term.
\begin{equation}
    \mathcal{L}_{SFT}(y,x) = -\alpha \log(\pi^*(y|x))
    \label{eq:sft}
\end{equation}

In order to introduce preferences more fine-grained than the sequence level to the model, we introduce non-term token masking into the loss. To do this, we construct a set of token indices, and compute 
$$\tilde{\pi}_\delta(y|x) = \sqrt[|\delta|]{\Pi_{i\in\delta}\log(\pi(y_i|y_{<i},x))}$$
with all $i\in\delta$ being indices into the sequence $y$ such that $y_i$ is part of a term.

We then create a masked variant of both the preference optimization and SFT losses, with masked preference optimization (mPO) being
\begin{multline}
    \mathcal{L}_{mPO}(x,y_w,y_l, \delta_w, \delta_l) = \\ sl_1\left(\log(\tilde{\pi}_{\delta_w}^*(y_w|x)) - \log(\!\tilde{\pi}_{\delta_l}^*(y_l|x)), \tfrac{1}{2\beta}\right)
    \label{eq:mpo}
\end{multline}
and masked supervised fine-tuning (mSFT) as
\begin{equation}
\mathcal{L}_{mSFT}(x, y_w, \delta_w) = -\alpha\log(\tilde{\pi}^*(y_w|x, \delta_w))
\label{eq:msft}
\end{equation}
These variants can be combined with each other such that we can perform preference optimization across entire sequences with extra emphasis on the terminology tokens. Or we could perform supervised fine-tuning on only terms.

We assign an indicator to each loss component to switch it on and off for different training runs. This results in our final loss function,
\begin{multline}
\mathcal{L}_{term} = \mathbb{1}_{PO}\mathcal{L}_{PO} + \mathbb{1}_{mPO}\mathcal{L}_{mPO} \\ + \alpha(\mathbb{1}_{SFT}\mathcal{L}_{SFT} + \mathbb{1}_{mSFT}\mathcal{L}_{mSFT})
\end{multline}
a combination of loss functions \eqref{eq:po}, \eqref{eq:sft}, \eqref{eq:mpo}, and \eqref{eq:msft}, which we evaluate with multiple settings to find the best for learning ambiguous terms.

\section{Experiments}

We perform two sets of experiments, a set of prompting experiments to serve as a baseline and fine-tuning experiments with variations of the $\mathcal{L}_{term}$ loss. For our prompting experiments, we use GPT 4.1\footnote{gpt-4.1-2025-04-14} from OpenAI. The prompting experiments were performed to see if commercial LLMs can perform the term disambiguation task without 
any training. We prompt with and without the term dictionary for the given source segment. Prompt templates for this experiment are available in Appendix \ref{app:prompt_templates}. Results for both the prompting and fine-tuning experiments are shown in section \ref{sec:results}, in Tables \ref{tab:baseline_results} and \ref{tab:term_results}, respectively.

\begin{table*}
    \centering
    \begin{tabular}{cccc}
    \toprule
        Model & ChrF& COMET & Term Accuracy    \\
       \midrule
    GPT 4.1 w Terms & 69.5 & 82.24 & 37.1 \\
    GPT 4.1 w/o Terms & 69.7 & 82.58 & 43.5 \\

    \bottomrule
    \end{tabular}
    \caption{Results from prompting experiments to serve as a baseline. GPT 4.1 refers to gpt-4.1-2025-04-14.}
    \label{tab:baseline_results}

\end{table*}

\subsection{Data}

We begin with a large corpus of post-edits on English to German machine translations produced by multiple NMT systems over multiple years. From our corpus of post-edits, we select examples of machine translations and post-edits containing ambiguous terminology with the method outlined in section \ref{terminology_matching} to create a training set of 123,518 examples. This data contains only examples where the machine translation contains a term translation in our dictionary, but the post-editor corrected it to a different translation from the dictionary. We additionally select 6000 examples for validation, and 2000 for testing. Our validation and test sets have evenly balanced terminology and non-terminology subsets.
Our training data contains 3579 unique source terms. On average, each ambiguous term in our term dictionary has 3.32 possible translations with a standard deviation of 1.85. The most extreme case was the term 'transfer' with 27 possible translations in our dictionary. In our test set, we find 335 unique source terms with 4.89 possible translations on average. Figure \ref{fig:term_histogram} in Appendix \ref{app:term_ambiguity} shows a histogram of the target term counts. The training data terminology has high coverage of the test set, with 97.8\% of terms in the test data also appearing in the training data. On our test set, we calculate the term accuracy that could be achieved by a random baseline as 24.9\%. The random baseline is simply a random choice over all possible term translations given the source term.



\subsection{Model Training}

\begin{table*}[h]
    \centering
    \begin{tabular}{cccccccccc}
    \toprule
       Init & Model & SFT & mSFT & PO & mPO & $\alpha$  & ChrF& COMET & Term Accuracy    \\
       \midrule
    Tower & Baseline & \cmark & \xmark & \xmark & \xmark & 1 & 73.5 & 83.14 & 53.7  \\
    Baseline & 1 & \xmark & \cmark & \xmark & \xmark & 1 & 72.3 & 82.74 & 56.1  \\
    Baseline & 2 & \cmark & \cmark & \xmark & \xmark & 1 & 73.2 & 83.01$^\dagger$ & 55.8   \\
    Baseline & 3 & \cmark & \xmark  & \cmark & \xmark & 1 & 72.2 & 82.65 & 54.6  \\
    Baseline & 4 & \xmark & \cmark & \xmark & \cmark & 10 & 72.0 & 82.62 & 55.9  \\
    Baseline & 5 & \cmark & \xmark  & \xmark & \cmark & 10 & 73.5 & 83.15$^\dagger$ & 55.6$^*$ \\
    Baseline & 6 & \cmark & \cmark  & \cmark & \cmark & 10 & 73.0 & 83.03$^\dagger$ & 56.3$^*$ \\
    \bottomrule
    \end{tabular}
    \caption{Results for rebooted experiments with Baseline trained only on the term data. COMET results marked with $\dagger$ have \textit{not} significantly changed from the baseline. Term Accuracy results marked with a $*$ \textit{have} significantly improved over the baseline.}
    \label{tab:term_results}

\end{table*}

We begin with Tower Base LLM 7B \citep{alves2024tower} and evaluate it without any additional training on our test set to establish its performance. Initially, it is translating 36.1\% of the terms correctly, achieving a COMET score of 78.72 and a ChrF score of 63.9. We then perform continued pre-training on the 123,518 English-German training examples to adapt it to our domain and to begin learning our terminology---the resulting model serves as our baseline LLM.

Hyperparameters for the continued pre-training step and further fine-tuning can be found in Appendix \ref{sec:hyperparameters}. For all of our trainings we perform full fine-tuning using Accelerate\footnote{\url{https://github.com/huggingface/accelerate}} with distributed data-parallelism. For our baseline model we use COMET score on the validation set as an early stopping mechanism. 

From this initial training, the model already learns our term dictionary quite well. 94.4\% of the time, it is able to correctly predict a valid term in our dictionary when faced with an ambiguous source term. However, that does not necessarily mean it is getting exactly the correct term. When we check for the exact term used in the test data, we find that only 53.7\% of the time is it picking the correct term translation. Additionally, it produces the exact same mistake that appears in the original machine translations 34.4\% of the time, suggesting that the knowledge-editing approach with preference optimization is appropriate. 

Once we have our baseline LLM, we perform fine-tuning for our term disambiguation task with various configurations of $\mathcal{L}_{term}$. Instead of using COMET as an early stopping criterion, we now use term accuracy only. 


We consider 6 different settings of $\mathcal{L}_{term}$. Setting 1 has only $mSFT$ and 2 has $SFT + mSFT$ enabled. 
Settings 3 through 6 then contain different combinations of preference optimization and supervised fine-tuning. 

Setting 3 examines $SFT + PO$, which is similar to the dCPO set-up of \citet{berger-etal-2024-post}. Setting 4 considers only the masked variants, so $mSFT + mPO$, to examine if only optimizing the term tokens suffices. Setting 5 uses $SFT + mPO$ to ensure that the entire post-edit remains likely while attempting preference optimization on terms only. Setting 6 combines all objectives $SFT + mSFT + PO + mPO$ to see if the objectives are complementary.


\subsection{Metrics}

Following the previous shared task for terminology translation at WMT \citep{semenov-etal-2023-findings}, we evaluate translations produced by our models with COMET \citep{rei-etal-2022-comet}, ChrF \citep{popovic-2015-chrf}, and term accuracy. 

The COMET model we use is based on the direct-assessment variant with references\footnote{\url{https://huggingface.co/Unbabel/wmt22-comet-da}} but fine-tuned on our own direct assessment data. \citet{zouhar-etal-2024-fine} show that fine-tuned neural metrics, such as COMET, show worse correlation with human judgements on out of domain data. We find that our fine-tuned COMET model has higher correlations with human judgements on our internal domains than the version released by \citet{rei-etal-2022-comet}.

In accordance with our data pre-processing, we also perform fuzzy matching with terminology to evaluate our term accuracy. We search for terminology in translation outputs with a threshold of 0.95 to allow for minor variations, such as different inflections for verb forms or plurals for nouns. If a match is found within this threshold, we count it towards our term accuracy.

\section{Results}

\label{sec:results}

In Table \ref{tab:baseline_results}, we report ChrF, COMET, and term accuracy for the GPT 4.1 prompting experiments. GPT 4.1, without the terminology, translates quite well, achieving a COMET score of 82.58. Without the term dictionary, it achieves a term accuracy of 43.5\%. 
If we consider any possible term translation and not only the correct term translation, then GPT 4.1 without the dictionary would achieve a term accuracy of 86.9\%. 
When given the term dictionary, its ability to use the correct term deteriorates, dropping to 37.1\%. 
However, if we consider again term accuracy over any possible term translation, then it would achieve a term accuracy of 98.2\%. 
This suggests that while prompting might be a straightforward way to use a dictionary, an LLM is not quite capable of disambiguating our terminology. This further motivates the use of a fine-tuning approach.

In Table \ref{tab:term_results} we report the term accuracy, ChrF and COMET scores for our all of our trained models. For our fine-tuning runs using $\mathcal{L}_{term}$, we report the setting of our indicators $\mathbb{1}$ with \cmark and \xmark.

Our baseline system achieves a term accuracy of 53.7\%, better than the 24.9\% we would expect from a random choice of term translations and the GPT 4.1 result of 43.5\%. This baseline is then used to initialize all following experiments in Table \ref{tab:term_results}. 

When fine-tuning our baseline model, we see that all settings of $\mathcal{L}_{term}$ positively affect term accuracy, with the best setting being all indicators set to $1$ (model 6). This achieves a term accuracy of 56.3\%. The combination of $\mathcal{L}_{SFT} + \mathcal{L}_{mSFT}$ also brings improvements and achieves a term accuracy of 55.6\%.
Settings 5 and 6 achieve significant improvements over the baseline, with $p < 0.05$ according to a pairwise approximate randomization significance testing \citep{riezler-maxwell-2005-pitfalls}. Although there are other results that achieve higher term accuracy, their lack of significance suggests that these results have higher variance.

Regarding machine translation quality, we see that settings 5 and 6 yield COMET scores that are not significantly different from the baseline.
Some training settings see degradation with regards to ChrF and COMET score while making gains in term accuracy, specifically settings 1, 3, and 4. This is seen starkly in settings that do not see the entire sequences but rather just the terms. Settings 1 and 4, consisting of $mSFT$ and $mSFT + mPO$ respectively, lose 1.2 and 1.5 ChrF points and 0.40 and 0.52 COMET points. This loss in COMET is a significant difference from the baseline model. This suggests that focusing only on terminology can cause the model to begin forgetting some more general translation abilities. However, settings 2, 5, and 6 do not see significant losses in terms of COMET score. Notably, all of these settings contain $\mathcal{L}_{SFT}$, indicating the continuing to train with SFT on the full sequence is necessary to mitigate losses while still gaining term accuracy. 



\section{Conclusion}

We introduced a preference optimization-based approach to improve terminology disambiguation in neural machine translation. By leveraging post-edited corrections of ambiguous terms, we trained models to better select the correct translation from multiple valid options.

Our best-performing model (setting 6: $\mathcal{L}_{SFT} + \mathcal{L}{mSFT} + \mathcal{L}_{PO} + \mathcal{L}{mPO}$ ) improved term accuracy from 53.7\% to 56.3\%. These gains come without significant losses on COMET, while models trained on term-focused losses alone saw larger degradation in terms of COMET and ChrF. This suggests that term-focused optimization can shift model behavior in ways that improve terminology but only insignificantly reduce overall fluency \citep{gisserot-boukhlef-etal-2024-preference}.

One hypothesis for the trade-offs is that semantically similar term translations, such as Übergabe and Übernahme for transfer, are likely close in embedding space, making it difficult for the model to separate them. If PO forces the model to distinguish between highly related terms, it may unintentionally alter how word choices are processed---leading to losses in general translation quality. Because of this, it is necessary to continue supervised fine-tuning on full sequences. 



\section{Limitations}

Limitations of our work include a broader evaluation on more language pairs. This shortcoming is due to the size of our post-editing data, which is largest for the English-German language pair, while post-edits for other language pairs are too small for training purposes. Furthermore, alternative preference optimization techniques, e.g., reinforce-style \cite{AhmadianETAL:24} with verifiable rewards on terminology fuzzy matches, similar to \citet{Lambert2024TLU3P} and \citet{rei2025towerbridginggeneralitytranslation}, could also be applied to this task and serve as a comparison point for our methods. 

\bibliography{main}

\newpage 

\appendix

\FloatBarrier

\section{Hyperparameters}
\label{sec:hyperparameters}

\begin{table}[H]
\centering
    \begin{tabular}{l l}
    \toprule
    Hyperparameter & Value \\
    \midrule
    Max Epochs & 20 \\
    Learning Rate & 1e-5\\
    Optimizer & AdamW \\
    Learning Rate Scheduler & Cosine \\
    Warm-up Ratio & 0.05 \\
    Effective Batch Size & 256 \\
    Max Gradient Norm & 10.0 \\
    Mixed Precision & \texttt{bfloat16} \\
    \midrule
    Early Stopping Criterion & Internal COMET \\
    Early Stopping Patience & 3 \\
    Early Stopping Epsilon & 0.00001 \\
    Evaluation Frequency & 1000 steps \\
    Max New Tokens & 64 \\
    \midrule
    Average Log-Probabilities & \texttt{True} \\
    Normalize Loss & \texttt{True} \\
    \bottomrule
    \end{tabular}
\caption{Hyperparameters for our baseline model with continued pre-training on all post-edit data.}
\label{tab:hyperparams_for_baseline}
\end{table}

\begin{table}[H]
\centering
    \begin{tabular}{l l}
    \toprule
    Hyperparameter & Value \\
    \midrule
    Max Epochs & 20 \\
    Learning Rate & 2e-6\\
    Optimizer & AdamW \\
    Learning Rate Scheduler & Cosine \\
    Warm-up Ratio & 0.05 \\
    Effective Batch Size & 256 \\
    Max Gradient Norm & 1.0 \\
    Mixed Precision & \texttt{bfloat16} \\
    \midrule
    Early Stopping Criterion & Term Accuracy \\
    Early Stopping Patience & 3 \\
    Early Stopping Epsilon & 0.00001 \\
    Evaluation Frequency & 250 steps \\
    Max New Tokens & 64 \\
    \midrule
    $\beta$ & $0.25$ \\
    Average Log-Probabilities & \texttt{True} \\
    Normalize Loss & \texttt{True} \\
    \bottomrule
    \end{tabular}
\caption{Hyperparameters for fine-tuning on terminology containing data subset}
\label{tab:hyperparams}
\end{table}


\FloatBarrier
\clearpage

\section{Term Ambiguity} 
\label{app:term_ambiguity}
\begin{figure*}[!htbp]
    \includegraphics[width=\textwidth]{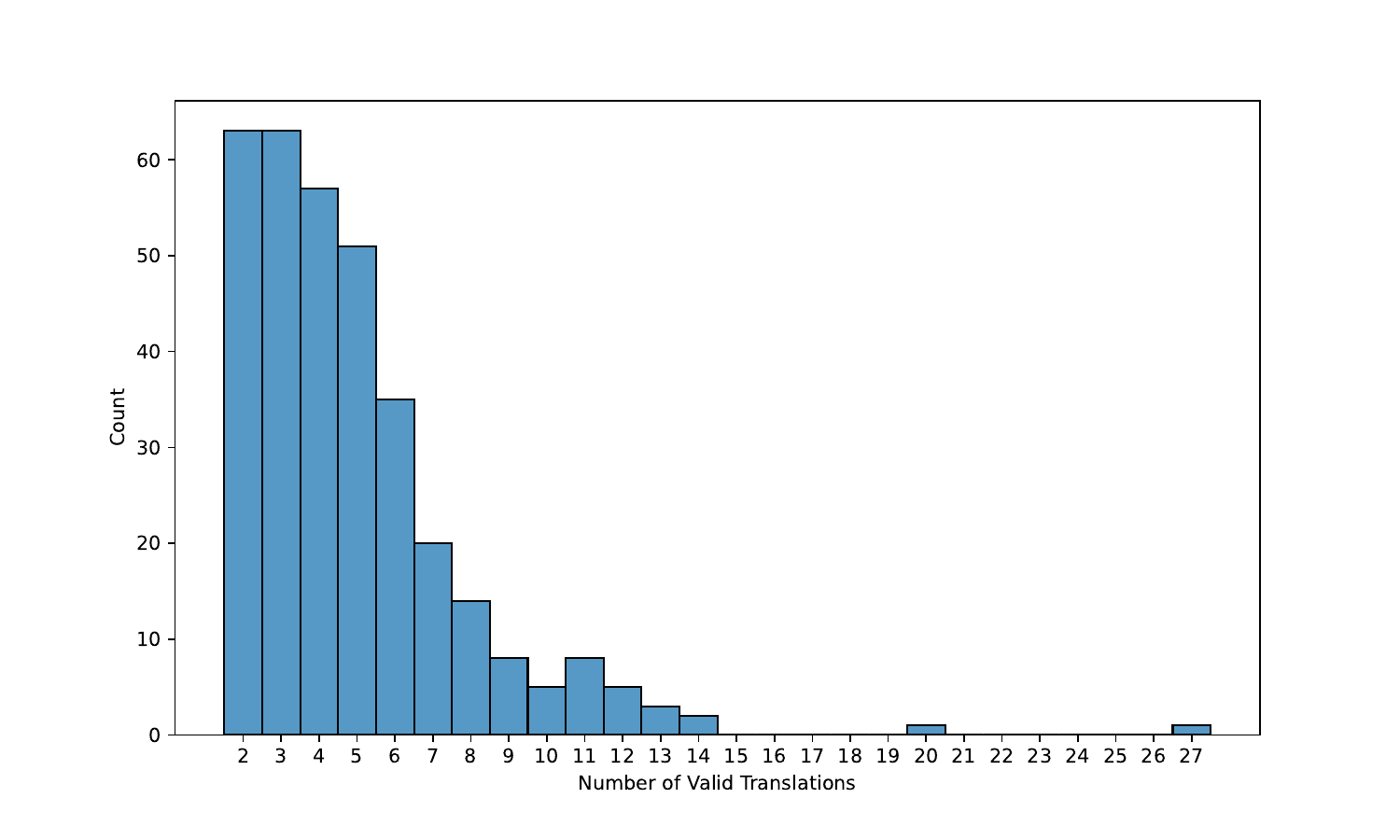}
    \caption{Here we show a histogram of the number of unique source terms in the test set for a given amount of valid translation terms. For the 335 unique source terms in the test set, on average they have 4.89 possible translations with a standard deviation of 2.97.} 
    \label{fig:term_histogram}

\end{figure*}

\FloatBarrier
\clearpage

\section{Sample dictionary entry}

\begin{figure*}[!htbp]
\texttt{transfer -> \{Versetzung, weitergeben, Warenbewegung, umlagern, Verlegung, übernehmen, Raumüberwindung, übertragen, Weiterleitung, Übertragung, Umbuchung, verlegen, Warenüberführung, Umladung, abführen, umbuchen, Versendung, Umlagerung, Übernahme, überleiten, Transfer, weiterleiten, Überführung, transferieren, Übergabe, Überleitung, Verfügung\}}
\caption{Mapping from a source term, 'transfer', to multiple valid target terms. Our term dictionary is a one-to-many mapping. Some target terms are also semantically overlapping, causing difficulties for the language model to produce the correct term translation}
\label{fig:term_translations}
\end{figure*}



\FloatBarrier
\clearpage
\section{Prompt Templates}

\label{app:prompt_templates}

\begin{figure*}[!hbp]
\small
\begin{verbatim}
[
  {
    "role": "system",
    "content": "You will be provided with a user input in English. Translate the text into German.
    
    The translation must satisfy the following terminology constraints:
    reference quantity - Bezugsmenge, Referenzmenge, Preiseinheit
    If more than one translation is possible for a given term, please select the best term.
    
    Only output the translated text, without any additional text."
  },
  {
    "role": "user",
    "content": "You can display the reference quantity and the simulation quantity side by side."
  }
]
\end{verbatim}
\hrule
\begin{verbatim}
[
  {
    "role": "system",
    "content": "You will be provided with a user input in English. Translate the text into German.
    
    Only output the translated text, without any additional text."
  },
  {
    "role": "user",
    "content": "You can display the reference quantity and the simulation quantity side by side."
  }
]
\end{verbatim}
\label{fig:sample_prompt}
\caption{Sample prompts for GPT 4.1. The upper prompt shows the added terminology constraints while the lower prompt is without constraints.}
\end{figure*}

\end{document}